\documentclass{article} 
\usepackage{nips12submit_e,times} \usepackage{graphicx}

\title{Multi-GPU Training of ConvNets}

\author{Omry Yadan \And Keith Adams \And 
 Yaniv Taigman  \And 
  Marc'Aurelio Ranzato \AND
 Facebook AI Group
  \\ \texttt{\{omry, kma, yaniv, ranzato\}@fb.com}}

\nipsfinalcopy 

\begin{document}

\maketitle


\begin{table}[b]
\caption{\footnotesize Comparison of different
  parallelization schemes. All models use a mini-batch size equal
  to 256 samples. The network is a convolutional network with the
  same structure and hyper-parameter setting as described by Krizhevsky et
al.~\cite{Krizhevsky:NIPS12} (with the only exception of the
mini-batch size). The task is classification on the
ImageNet 2012 datasset~\cite{imagenet}. All GPUs are NVIDIA TITANs
with 6GB of RAM and they all reside on the same server.}
\label{tab:comparison}
\centering
\begin{tabular}{c||c}
{\bf Configuration} & {\bf Time to complete 100 epochs}
\tabularnewline 
\hline
\hline 
1 GPU & 10.5 days
\tabularnewline 
\hline 
2 GPUs Model parallelism & 6.6 days
\tabularnewline 
\hline 2 GPUs Data parallelism & 7 days
\tabularnewline 
\hline 4 GPUs Data parallelism & 7.2 days
\tabularnewline 
\hline 
4 GPUs model + data parallelism & 4.8 days
\end{tabular}
\end{table}

Convolutional neural networks~\cite{lecun-98} have proven useful
in many domains, including computer vision
~\cite{Krizhevsky:NIPS12, nipsobjdetection13, farabet-scene-parsing13},
audio processing~\cite{convnet-speech12, convnet-speech13} and
natural language processing~\cite{convnet-collobert11}. 
These powerful models come at great cost in training time,
however. Currently, long training periods make experimentation difficult and time consuming.

In this work, we consider a standard
architecture~\cite{Krizhevsky:NIPS12} trained on the Imagenet
dataset~\cite{imagenet} for classification and investigate methods
to speed convergence by parallelizing training across multiple
GPUs. In this work, we used up to 4 NVIDIA TITAN GPUs with 6GB of RAM. 
While our experiments are performed on a single server, our GPUs
have disjoint memory spaces, and just as in the distributed
setting, communication overheads are an important
consideration. Unlike previous work~\cite{dean12, asgd-gpu,
  pipelined-gpu}, we do not aim 
to improve the underlying optimization algorithm. Instead, we isolate
the impact of parallelism, while using standard supervised
back-propagation and {\em synchronous} mini-batch stochastic gradient
descent. 

\begin{figure}[th]
\begin{centering}
\includegraphics[width=0.2\linewidth]{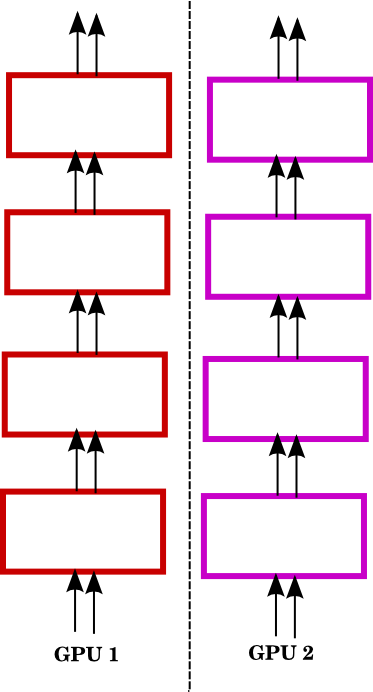}
\par\end{centering}

\caption{{\footnotesize{Diagram of a generic deep network using
      two GPUs ({\em data parallelism}). Each GPU computes errors
      and gradients for half of the samples in the
      mini-batch. Parameters and gradients are communicated across
      GPUs using PCI-e. The layers computed on a
  GPU all share the same color in the diagram.}}}

{\footnotesize{\label{fig:outline-dataparallel} }}
\end{figure}

We consider two basic approaches: data and model
parallelism~\cite{dean12}. In data parallelism, the mini-batch is
split across several GPUs as shown in
fig.~\ref{fig:outline-dataparallel}.  Each GPU is responsible for
computing gradients with respect to all model parameters, but does
so using a subset of the samples in the mini-batch. This is the
most straightforward parallelization method, but it requires
considerable communication between GPUs, since each GPU must
communicate both gradients and parameter values on every update
step. Also, each GPU must use a large number of samples to
effectively utilize the highly parallel device; thus, the
mini-batch size effectively gets multiplied by the number of GPUs,
hampering convergence. In our implementation, we find a speed-up
of 1.5 times moving from 1 to 2 GPUs in the data parallel
framework (when using 2 GPUs, each gets assigned a mini-batch of
size 128). This experiment used the same architecture,
network set up and dataset described in Krizhevsky et
al.~\cite{Krizhevsky:NIPS12}.

\begin{figure}[th]
\begin{centering}
\includegraphics[width=0.4\linewidth]{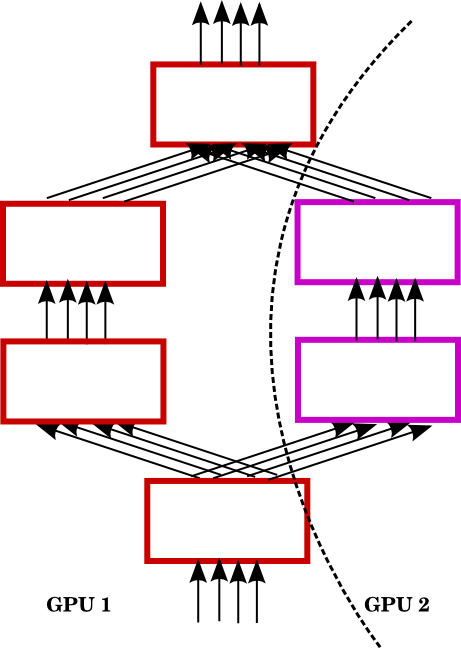}
\par\end{centering}

\caption{{\footnotesize{Diagram of a generic deep network using
      two GPUs ({\em model parallelism}). The architecture is
      split into two columns which makes easier to split
      computation across the two GPUs.}}}

{\footnotesize{\label{fig:outline-modelparallel} }}
\end{figure}

Model parallelism, on the other hand, consists of
splitting an individual network's computation across
multiple GPUs~\cite{dean12,coates13}. An example is shown in
fig.~\ref{fig:outline-modelparallel}. For instance, a
convolutional layer with N filters can be run on two GPUs,
each of which convolves its input with N/2 filters. 
In their seminal work, Krizhevsky et
al.~\cite{Krizhevsky:NIPS12} further customized the architecture
of the network to better leverage model parallelism: the
architecture consists of two ``columns''
each allocated on one GPU. Columns have cross connections only at
one intermediate layer and at the very top fully connected
layers. While model parallelism is more difficult to implement, it
has two potential advantages relative to data parallelism. First, it may requires less communication
bandwidth when the cross connections involve small intermediate
feature maps. Second, it allows the instantiation of models that
are too big for a single GPUs memory. Our implementation of
model parallelism, replicating prior work by Krizhevsky  et
al.~\cite{Krizhevsky:NIPS12}, achieved a speed-up of 1.6 times
moving from 1 to 2 GPUs. 

\begin{figure}[th]
\begin{centering}
\includegraphics[width=0.8\linewidth]{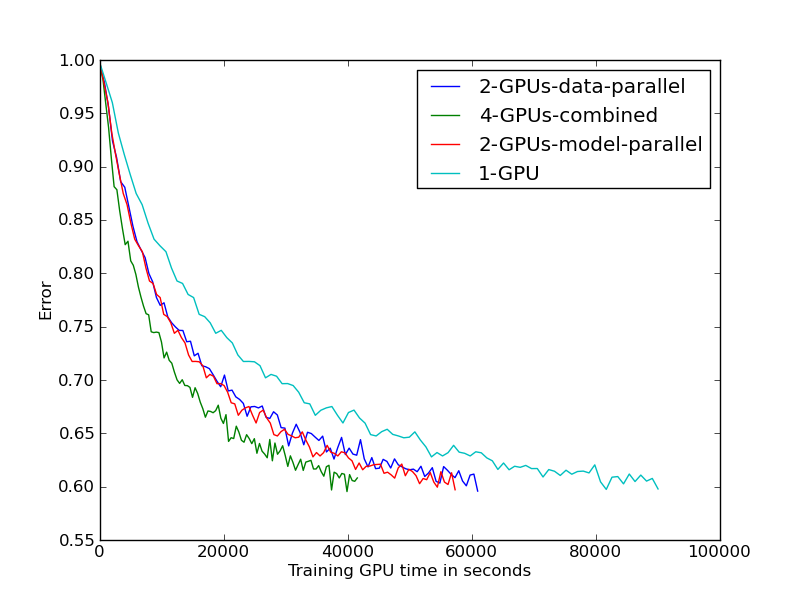}
\par\end{centering}
\caption{{\footnotesize{Test error on the ImageNet dataset as a function of time using different forms
      of parallelism.  On the y-axis we report the error rate
      of the test set, on the x-axis time in seconds. All
      experiments used the same mini-batch size (256) and ran for 100
      epochs (here showing only the first 10 for clarity of
      visualization) with the same architecture and the same
      hyper-parameter setting as in Krizhevsky  et
al.~\cite{Krizhevsky:NIPS12}. If plotted against number of weight
      updates, all these curves would almost perfectly
      coincide.}}}
{\footnotesize{\label{fig:comparison-4gpus_crop} }}
\end{figure}

\begin{figure}[th]
\begin{centering}
\includegraphics[width=0.8\linewidth]{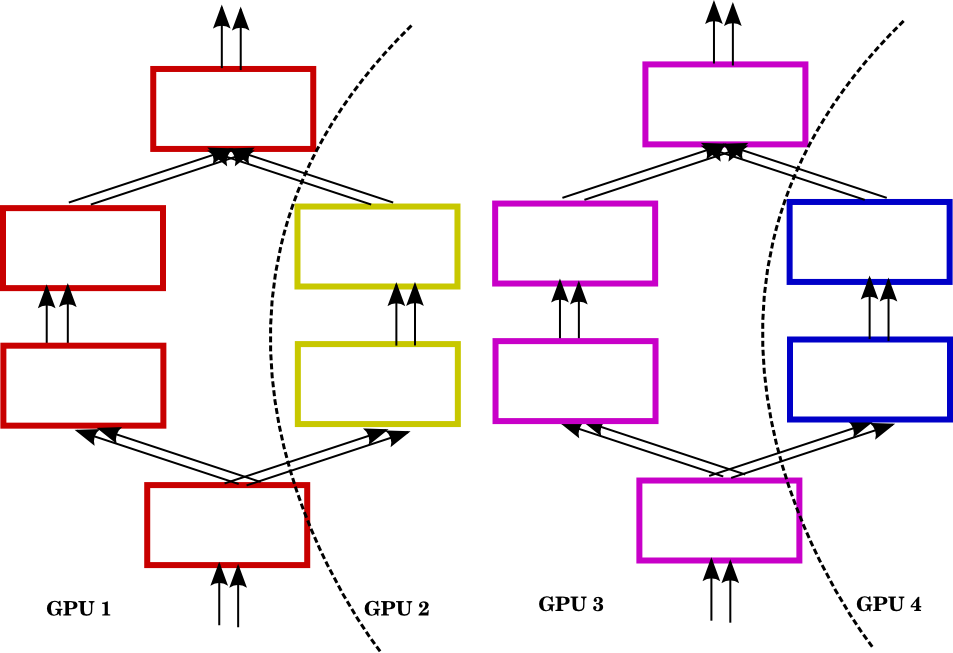}
\par\end{centering}

\caption{{\footnotesize{Example of how model and data parallelism
      can be combined in order to make effective use of 4 GPUs.}}}

{\footnotesize{\label{fig:outline-datamodel-parallel} }}
\end{figure}

Data and model parallelism can also be hybridized, as shown in
fig.~\ref{fig:outline-datamodel-parallel}. We consider using 4
GPUs in three possible configurations: all model parallelism, all
data parallelism, and a hybrid of model and data parallelism. In
our experiment, we find that the hybrid approach yields the
fastest convergence. The hybrid approach on 4 GPUs achieves a
speed-up of 2.2 times compared to 1 GPU. More detailed results are
shown in tab.~\ref{tab:comparison}. The convergence curves
comparing the most interesting configurations are shown in
fig.~\ref{fig:comparison-4gpus_crop}. 

In general, not all configurations could be explored.
For instance, on a single NVIDIA TITAN GPU with 6GB of RAM we are unable to fit
mini-batches larger than 256 samples. 
On the other hand, we find mini-batches of 64 or fewer samples
under-utilize the GPUs cores. This can be seen in
tab.~\ref{tab:comparison} which reports the timing of the data
parallelism approach using 4 GPUs.

These preliminary results show promising speed-up factors by
employing a hybrid parallelization strategy. In
the future, we plan to extend this work to parallelization across
servers by combining data and model parallelism with recent
advances in asynchronous optimization methods and local learning
algorithms. \\

\begin{figure}[th]
\begin{centering}
\includegraphics[width=0.1\linewidth]{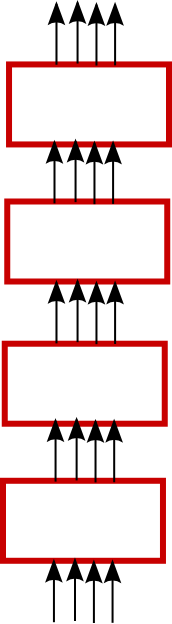}
\par\end{centering}

\caption{{\footnotesize{Diagram of a generic deep network. The
      number of arrows is proportional to the size of the
      mini-batch.}}}

{\footnotesize{\label{fig:outline-single} }}
\end{figure}

{\small{\bibliographystyle{unsrt}
    \bibliography{parallelGPU_iclr2014} }}
\end{document}